\documentclass[letterpaper, 10pt, conference]{ieeeconf}      

\IEEEoverridecommandlockouts                              

\usepackage{array,multirow,graphicx}
\usepackage{hyperref}
\usepackage{amsmath}
\usepackage{caption}
\usepackage{amssymb}
\usepackage{siunitx}
\usepackage{caption}
\usepackage{subcaption}
\usepackage{tabularx}
\usepackage{url}
\usepackage{float}
\usepackage{pgf-umlsd}
\usepgflibrary{arrows} 
\usepackage{verbatim}
\usepackage{xcolor}
\usepackage{hyperref}
\usepackage{footmisc}
\usepackage{multirow}
\usepackage{xspace}

\newcommand{\argmax}{\operatornamewithlimits{argmax}}

\graphicspath{ {images/} }

\usepackage{multirow}

\title{Context-aware robot control using gesture episodes}

\author{
Petr Vanc$^{1}$
\and
Jan Kristof Behrens$^{1}$
\and 
Karla Stepanova$^1$
\thanks{$^{1}$Czech Technical University in Prague,
Czech Institute of Informatics, Robotics, and Cybernetics, \texttt{petr.vanc@cvut.cz}, \texttt{jan.kristof.behrens@cvut.cz}, \texttt{karla.stepanova@cvut.cz}}
}

\begin{document}

\maketitle
\thispagestyle{empty}
\pagestyle{empty}

\begin{abstract}
Collaborative robots became a popular tool for
increasing productivity in partly automated manufacturing
plants. Intuitive robot teaching methods are required to quickly
and flexibly adapt the robot programs to new tasks. Gestures
have an essential role in human communication. However,
in human-robot-interaction scenarios, gesture-based user
interfaces are so far used rarely, and if they employ a one-to-one
mapping of gestures to robot control variables. In this paper, we
propose a method that infers the user’s intent based on gesture
episodes, the context of the situation, and common sense. The approach is evaluated in a simulated table-top manipulation setting.
We conduct deterministic experiments with simulated users and
show that the system can even handle personal preferences of
each user.

\end{abstract}


\section{Introduction}
\label{sec:intro}

Robots are complex machines with unintuitive kinematics and operational constraints. When a worker wants to instruct a robot to do a set of manipulations, it is more natural to convey the high-level manipulation goal (intent) than to program low-level robot actuations. Gestures are a natural part of human communication and have been used to communicate with robots since the late 1990s when the recognition of six different arm gestures was used to control a wheeled
robot~\cite{kortenkamp1996recognizing}. In contrast to the natural use of gestures in social situations, gestures as part of a user interface (UI) are considered \textit{functional}~\cite{carfi2021gesture}. Most of the current works focus on 1-to-1 mapping between gestures
and controlled variables~(e.g., \cite{coronado2017gesture,carfi2018online,Vanc_Stepanova_Behrens}), thus heavily restricting expressiveness of the gestures.

This paper proposes a system that allows the user to control a robot by communicating target actions, objects, and other parameters via gestures. Simultaneously, it leaves the robot the autonomy to execute the high-level intent by a suitable set of low-level robot actions, e.g., generated by a behavior tree \cite{Colledanchise_Ogren_2018}. We utilize the Leap Motion sensor \cite{Weichert_Bachmann_Rudak_Fisseler_2013} to detect and track the user's hand-bone structure. We do not impose any assumptions on hand gestures or their exact meaning, enabling the user to define new gestures via demonstration. The detected gestures (expressing a human intent) are fed to a probabilistic neural network for classification (see~Fig.\ref{fig:model_graph_01}). The use of the gestures might vary from user to user and based on the scene or manipulated object. For example, a gesture to open a drawer might be a pulling motion, while the opening of a box will rather be a lifting motion. We show the importance of including the context of the situation when inferring the user's intents on a set of artificial datasets. 

We demonstrate our system in a simulated table-top manipulation scenario with 7DoF robot manipulator (see~Fig.\ref{fig:cbgo_init_scene_random}). Let us assume the user intends to tidy up a cup in a drawer. She focuses her eyes on the drawer, then makes a gesture swipe down to signal the object's placing. In this way, the intent is assembled as the accumulated desired change in the world state. The system accumulates all gesture detection data over the episode and compresses it into an observation vector. This observation vector and the world state are then used to infer the user's intent.

The contributions of this paper are
\begin{itemize}
\item a flexible gesture recognition framework that also supports gesture combinations (for improved expressiveness) and context-aware gesture interpretation,
\item an approach to learn such a context-aware gesture-intent mapping from sampled data, and
\item a system implementation that grants the robot more autonomy and thereby increases the robustness of the gesture control system (see code and data at \href{https://github.com/imitrob/context-based-gesture-operation}{github.com/imitrob/context-based-gesture-operation})
\end{itemize}

\section{Related Work}\label{sec:related}


Based on the taxonomy introduced in the survey paper~\cite{carfi2021gesture}, we focus on functional static and dynamic manipulative and control hand gestures. 
The proposed system enables the definition and usage of both semaphoric gestures associated with a command and manipulative gestures that map the gesture movement to the location and pose of the object. 

Many works focus on developing better sensors enabling gesture-recognition such as wearable~\cite{kralik2021waveglove} or contact-less sensors~\cite{stetco2020gesture}. The development of better methods to detect individual gestures~\cite{coronado2017gesture,carfi2018online} and human activities provides important key components for HRI systems. 
Several papers utilize the Leap~Motion controller \cite{Leap_Motion_Controller} to detect hand gestures. Methods to learn and detect the gestures include deterministic learning \cite{Zeng_Wang_Wang_2018}, support vector machines (SVM) with custom combinations of features \cite{Du_Liu_Feng_Chen_Wu_2017}, or recurrent neural networks (RNN) \cite{Avola_Bernardi_Cinque_Foresti_Massaroni_2019}. \cite{Marin_2016} introduces hand gesture recognition using a Leap Motion sensor and Kinect depth camera.
These works typically focus on 1-to-1 mapping between gestures and controlled variable~\cite{coronado2017gesture}. Instead, we aim for more robot autonomy and a more expressive gesture language through a combination of context-dependent gestures.

 Designing a good UI is an art that is approached very systematically in \cite{coronado2017gesture}, where arm gestures detected by an inertial sensor are used to control a robot's functional states (e.g., off, idle, and moving) as well as the specific motion when the robot is moving. While we are not concentrating on making a single system as user-friendly as possible, we introduce concepts that let users operate systems in a personalized manner.

\cite{Iengo_Rossi_Staffa_Finzi_2014} introduces a system that detects gestures in real-time from a Kinect RGBD camera using ad-hoc Hidden Markov Models. Our system works as well in real-time. Our gesture detection using probabilistic neural networks also needs only a few sample demonstrations to learn new gestures. In contrast, we deal with a more complex HRI use case where the robot has more autonomy.
\cite{Cicirelli_Attolico_Guaragnella_DOrazio_2015} use Kinect RGBD camera and OpenNI to extract joint angles as features. A neural network classifier allows real-time gesture detection with a sliding window approach. However, only three joint angles were considered in the feature vector, and gestures have a fixed length of 2 seconds (60 frames). In this work, we represent dynamic gestures as Probabilistic Motion Primitives \cite{Paraschos_Daniel_Peters_Neumann_2013} and compare new data via dynamic time warping against the library of dynamic gestures. We also add context and user-dependent gesture selection on top. 

\section{Problem formalization}
\label{sec:problem_formalization}



\noindent Taking into account the current state of the world, the user expresses his desired intent (the desired change in the world) by hand and eye movements (i.e., gesture and focus point can be observed). The gesture-based system should be able given the observation $\textbf{o}$ to determine the user's intent $\textbf{i}$ and transform it to a sequence of robot actions $\textbf{a}$. In other words, the task is to learn from a set of observations the mapping $\mathcal{M}$ between the observations and the intents and the mapping $\mathcal{A}$ between the intent $\mathbf{i}$ and the robotic action sequence $\mathbf{a}$. Note that individual users might use different gestures in different contexts. Therefore the mapping has to take into account all the observation variables.

\textit{Observation} tuple $\textbf{o}=[\textbf{h}, \textbf{f}, \textbf{s} , u]$ contains time series of hand features $\textbf{h}$ and focus point \textbf{f}, accompanied by the context of the situation -- i.e., the scene description $\textbf{s}$ and the user description $u$.
\textit{Scene} $\textbf{s}$ expresses the state of the system. It includes information about the position, type, and state of the objects in the scene, the position and state of the end-effector, and possibly other features. \textit{User description} can be just identification of the user or additional parameters describing the user (e.g., emotional state, nationality, etc.). 

Given the time series of \textit{hand movement features} $\textbf{h}$, set of \textit{available gestures} $\mathbf{G}$ and the trained \textit{classifier} $\mathcal{G}$, the probability of individual \textit{gestures} $\mathbf{g}$ can be determined:
\begin{equation}
\label{eq:gestures}
\textbf{g} = \mathcal{G}(\textbf{h}, \mathbf{G})
\end{equation}
In this regard, the gesture is defined as a specific pattern performed by a user to affect the behavior of an intelligent system (see~\cite{carfi2021gesture}) -- we distinguish static (pose-based) or dynamic (motion-based) gestures. Each gesture is expressed by a model trained on a set of demonstrations. The gesture set can be arbitrarily extended. Note that each user might train their own set of gestures or use the general gestures available in the system.

We define \textit{intent} in the given context $\textbf{i}$ as a tuple of three variables: target action $ta$, target object $to$, and target metric $tm$ (selected based on their probability). 


\begin{equation}
\mathbf{i} = (\underbrace{argmax(\textbf{i}^{probs}_{ta})}_{ta}, \underbrace{argmax(\textbf{i}^{probs}_{to})}_{to}, tm).    
\end{equation}
The probabilities of the intended actions $\textbf{i}^{probs}_{ta}$ and target objects $\textbf{i}^{probs}_{to}$ in the given context are determined by applying a mapping on observation tuple:
$\textbf{i}^{probs}_{ta} = \mathcal{M}_a(\textbf{o})$,
$\textbf{i}^{probs}_{to} = \mathcal{M}_o(\textbf{o})$.
\noindent Target action $ta$ represents the action that the user wants to perform (e.g., open, push, move into, etc.), target object $to$ denotes an object instance from the scene to act on. Target metric $tm$ is, in general, a vector of $m$ metric parameters for the given actions. In this paper, we selected only actions with one metric parameter. Therefore we do not solve the the general assignment problem of values in $tm$ to target action parameters in this paper. The system selects the most probable feasible intent over a threshold. 

From the intent $\mathbf{i}$, the \textit{robotic action sequence} $\textbf{a}$ with specified parameters is computed, taking into account the set of actions within the action space $A$: $\textbf{a} = \mathcal{A}(\textbf{i})$.
We consider robotic actions in this regard as purposeful manipulation of the world by the robot. A sequence of actions is needed when the target action has unsatisfied preconditions. $\mathcal{A}$ can be implemented in general using task planning, but we opted for a solution based on a behavior tree that handles the execution reactively, including error recovery strategies.
We consider the following three cases. The intent can either correspond to: 1) a single robot action (e.g., pour water from the cup); 2) a single robot action with specific parameters (e.g., rotate the cup by 5 degrees); or 3) a sequence of robotic actions (e.g., put the cup to the drawer corresponds to a sequence of actions: get close to the drawer, open gripper, move gripper away).




\section{Materials and Methods}
\label{sec:methods}

\begin{figure}[htbp]
  \centering
  \includegraphics[width = 0.3\textwidth]{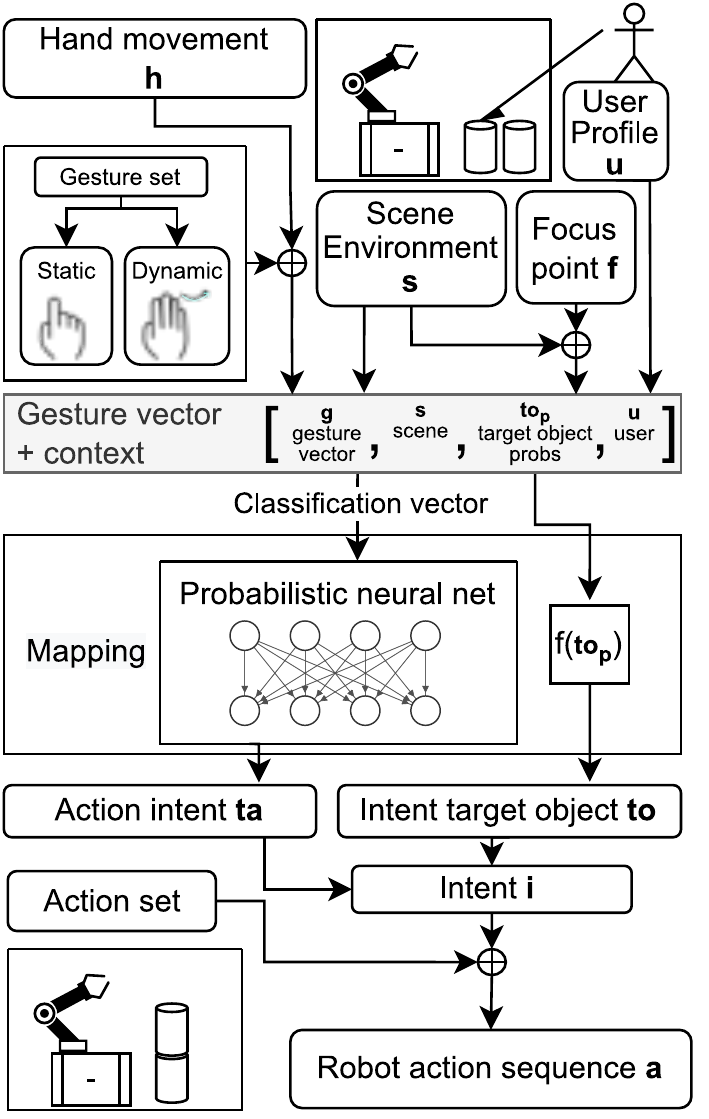}
  \caption{The diagram of the proposed system shows the whole pipeline from hand observations to robotic actions.}
  \label{fig:cbgo_diagram}
\end{figure}

The proposed system can be divided into three parts. The first handles the real-time gesture recognition task $\mathcal{G}$ (Fig.:~\ref{fig:cbgo_diagram}$^{\text{upper}}_{\text{part}}$) from hand observations $\textbf{h}$, the second uses the gesture classification vector $\textbf{g}$ and infers the user's intent $\textbf{i}$ (using mapping $\mathcal{M}$), and finally the third part realizes robotic action sequence $\textbf{a}$ for the given intent. 

\subsection{Gesture classification} 
\label{sec:gesture_classification}

We distinguish static and dynamic gestures in the proposed setup, classified by separate classifiers. 
A set of features describing the hand movement are computed from the data acquired from the Leap~Motion sensor~\cite{Weichert_Bachmann_Rudak_Fisseler_2013} (i.e., hand bone structure and hand position). 


For the gesture classification task  (see Eq.~\ref{eq:gestures})
we utilize our teleoperation gesture toolbox (see \href{https://github.com/imitrob/teleop_gesture_toolbox}{\url{github.com/imitrob/teleop\_gesture\_toolbox}}, \cite{Vanc_Stepanova_Behrens}).
The gesture sets (static and dynamic variants) can be adapted to the use case and user preferences, including learning new gestures directly from the user or adjusting the mapping between the existing gestures and the desired action. The system enables one to set the mapping between gestures and actions directly in GUI or learn it from observations.

\subsubsection{Static gestures}

Static gestures have a constant stroke phase. The following hand-crafted feature set is analyzed in every frame: 
1)~distances between fingertip positions ($\mathbf{x}_{f1},\hdots, \mathbf{x}_{f5}$) (including the distances to the palm center ($\mathbf{x}_{palm}$)), and 2)~joint angles between hand bones ($\alpha_1,\hdots,\alpha_N$) that are computed as angles between direction vectors (obtained from differences between individual bone positions). A total amount of 57 features are used:
\begin{equation}
    \mathbf{h} = [\alpha_1,...,\alpha_{42}, ||\mathbf{x}_{f1}, \mathbf{x}_{palm}||, ||\mathbf{x}_{f1}, \mathbf{x}_{f2}||, \hdots,||\mathbf{x}_{f4}, \mathbf{x}_{f5}||]
\end{equation}



The most accurate detection method for static gestures has proven to be the probabilistic neural network model (see Sec.~\ref{sec:nn_method}). 
Other tested methods included deterministic neural networks and a hand-crafted approach based on hand-picked thresholds. We made empirical and classification tests for various combinations of the features. In our setup, up to 10 static gestures were considered.

\subsubsection{Dynamic gestures}
\label{sec:methods_dynamic_gestures}
Compared to static gestures, dynamic gestures are characterized by a movement, i.e., the stroke phase is a trajectory. For simplicity, we represent in our experiments dynamic gestures only by the Cartesian position of the palm center in each time step $i$ ($i\in\{1,\hdots,N\}$): 

\begin{equation}
    \mathbf{h} = [x_1, y_1, z_1,\hdots, x_N, y_N, z_N]
\end{equation}

The used Leap Motion sensor records with up to $100$~Hz. According to~\cite{forbes_efficient_2005} it is possible to reduce measurement frequency to $f=\SI{20}{Hz}$ without any significant loss. 

Each dynamic gesture is learned from a set of observations and represented as a probabilistic motion primitive~\cite{Paraschos_Daniel_Peters_Neumann_2013}. For the classification task, the Dynamic Time Warping (DTW)~\cite{muller_dynamic_2007}~\cite{Salvador_Chan} is utilized, as it can deal with different motion speeds and time deformations and is very computational efficient (see~\cite{Vanc_Stepanova_Behrens} for the details of the method).

\subsubsection{Auxilary parameter extraction}
In addition to the semaphoric meaning of most gestures, we extract metric features from the hands that can be used as parameters for robotic actions. For example, distances can easily be demonstrated as the distance between the thumb and the index finger. The information on which types of parameters are required for a particular action is drawn from the action signature. Not-needed parameters are ignored, and missing parameters are filled with default values.

\subsection{Episode}

We consider an \textit{episode} as the time window in which the user specifies the intent using one or more gestures. In this paper, we assume that an episode begins with the detection of hands and ends when they disappear. To improve the ergonomy and stability of the system, 
we run the gesture recognizers on the movements (so-called \textit{stroke}) that started in the center of the detection area. This allows the user to always return to the center for the next gesture without accidentally triggering other gestures and avoid undesired \textit{preparation} and \textit{retraction} moves. The notion of episode enables us to collect observations over time until we accumulate enough evidence to infer an actionable intent.

The data are continuously fed to the circular buffer for the gesture recognition task. The parallel recognition of static and dynamic gestures is running (see Sec.~\ref{sec:gesture_classification}), accumulating the evidence for individual gestures. Therefore, a set of gestures passing the evidence threshold $E_t$ might be recognized within one episode (see sample episode evaluation Fig.:~\ref{fig:episode_graph_01}, where gestures "point" and "swipe down" were recognized). Anytime one of the gestures passes the evidence threshold, the probability vector for all gestures is saved. The set of the saved probability vectors within one episode is combined into one gesture vector $\mathbf{g}$, that gives information about the maximum probabilities of each gesture in the given episode: 



\begin{equation}
        \textbf{g} = max_i \left( \begin{bmatrix}g_{1,1} \\ g_{2,1} \\ \hdots \\ g_{G,1} \end{bmatrix} , \hdots, \begin{bmatrix} g_{1,i} \\ g_{2,i} \\ \hdots \\ g_{G,i} \end{bmatrix} , 
        \hdots , \begin{bmatrix} g_{1,C} \\ g_{2,C} \\ \hdots \\ g_{G,C} \end{bmatrix} \right)
\end{equation}

where $G$ is the number of considered gestures, and $C$ is the number of triggered gesture observations in the given episode.  
The gesture vector $\mathbf{g}$ is one of the inputs to the intent recognition system.

\subsection{Mapping represented by Probabilistic Neural Network}
\label{sec:nn_method}

\begin{figure}[htbp]
  \centering
  \includegraphics[width=0.13\textwidth, trim={2cm 1.5cm 2cm 1cm} ]{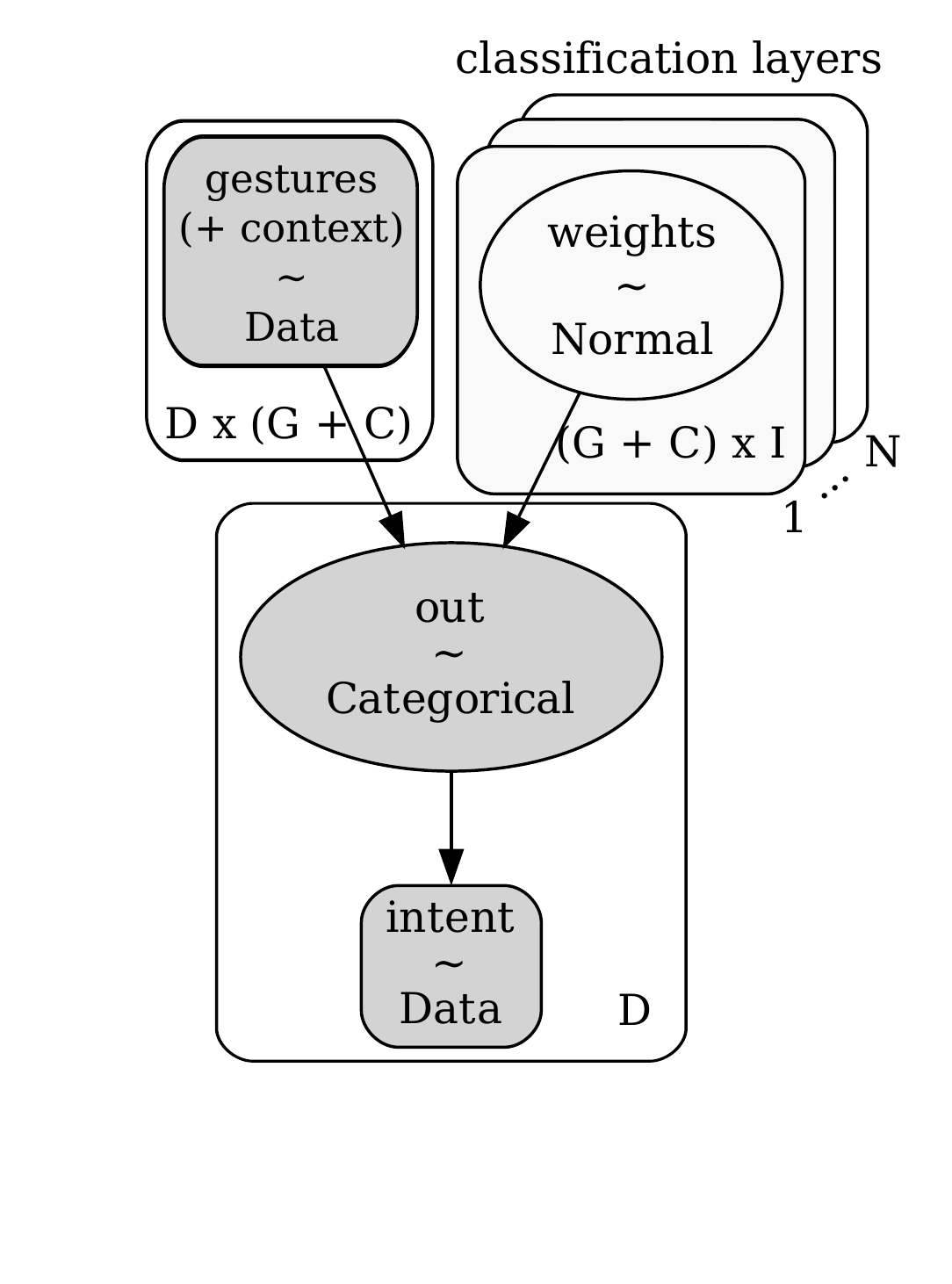}
  \caption{Graphical model of Probabilistic neural network training for the mapping task $\mathcal{M}$ (Gestures + Context $\rightarrow$ Intent), $D$ is \# of dataset samples, $G$ is \# of gestures, $C$ context observations, $I$ is user intent probabilities, and $N$ is \# of classification layers. \textit{Normal} and \textit{Categorical} tags represent type of probability distribution. \textit{Categorical} type represents discrete probabilistic distribution (see Sec.:~\ref{sec:nn_method}).}
  \label{fig:model_graph_01}
\end{figure}

We approach the problem of inferring user intent $\textbf{i}$ for a given set of gestures $\textbf{g}$ as a classification problem. We train a probabilistic neural network (1 to 3 fully connected hidden layers) in a supervised manner using data from sampled scenes with generated intents and gestures. To reflect the context-dependent usage of the gestures, we take as an input to the network a combination of gesture vector, user id, focus point, and scene object state and type (see Fig.~\ref{fig:model_graph_01}). The output is a categorical distribution with probability density function $f(i | \mathbf{p}) = p_i, \forall$ intents $i$, $\mathbf{p}$ is a vector of probabilities of individual intents ($\sum_i{p_i} = 1$, $p_i>0 \forall i$).

The forward loop of the network can be written as: 


\begin{equation}
        Y = sigmoid(\tanh(\mathbf{X} \cdot \textbf{w}_1) \cdot \textbf{w}_2),
\end{equation}
where $\mathbf{X}$ is input observations vector, $\textbf{w}_1$ and $\textbf{w}_2$ are classification weights of the length $O \times 25$ and $25 \times I$, respectively. $O$ is the length of observation vector (gesture vector + context) $O = G + C$, I is the amount of possible gestures/intents. The $\mathbf{Y}$ is a vector of $I$ intent/gesture probabilities. For the classification result, the selected class is chosen as the one with the maximum probability $c = \argmax_I \mathbf{Y}$.

The automatic differentiation variational inference (ADVI) is used as a fitting algorithm \cite{kucukelbir_automatic_2016}. Kullback-Leibler divergence is used as a loss function. The function callback is called every 2000th epoch and makes a few samples drawn on the test data. In this way, we can estimate the local maximum and avoid overfitting.
The network is build on PyMC framework \cite{emaasit_pymc-learn_2018}.

\subsection{Robotic action sequence generation}

\begin{figure}[htbp]
  \centering
  \includegraphics[width=0.35\textwidth]{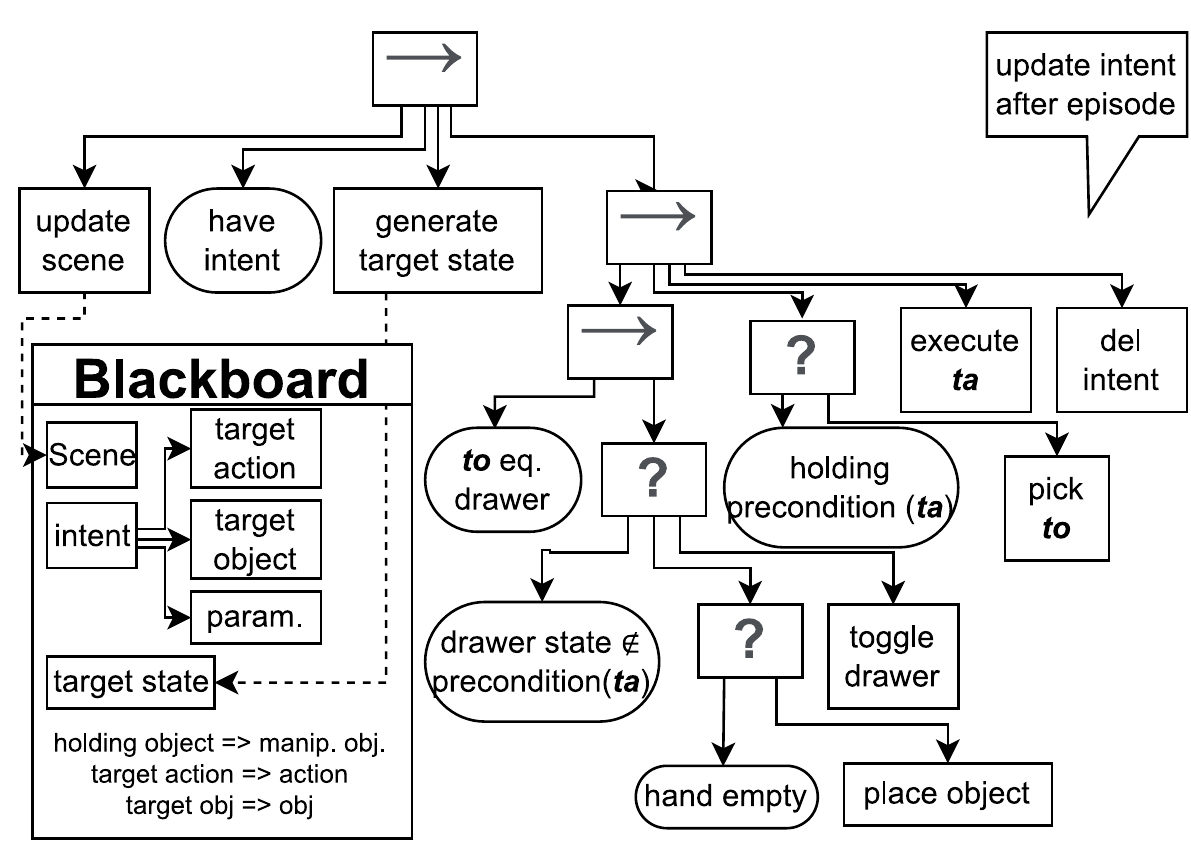}
  \caption{Construction of action sequence $\textbf{a}$ from intent $\textbf{i}$ using a Behavior tree approach.}
  \label{fig:cbgo_behavior_tree_example}
\end{figure}

The generation of robotic motions for an intent in the form of a \textit{target action}, a \textit{target objects}, and (optionally) a set of metric parameters utilizes a behavior tree \cite{Colledanchise_Ogren_2018}. The behavior tree encodes a high-level policy and is parameterized by the intent as a goal. An intent \textit{(put into, drawer)} while holding a \textit{cup} requires opening the drawer (if it is closed) and placing the cup inside the drawer. The behavior tree automatically triggers the opening action that the users rarely command. Hence, it is necessary to deal with incomplete user specifications. 
Fig.~\ref{fig:cbgo_behavior_tree_example} shows a visual representation of a behavior tree for our example domain.








\section{Experimental setup and dataset generation}
\label{sec:setup}

\subsection{Robotic environment and gesture classification}

\begin{figure}[htbp]
  \centering
  \includegraphics[width=0.23\textwidth, trim={1cm 2cm 1cm 2cm} ]{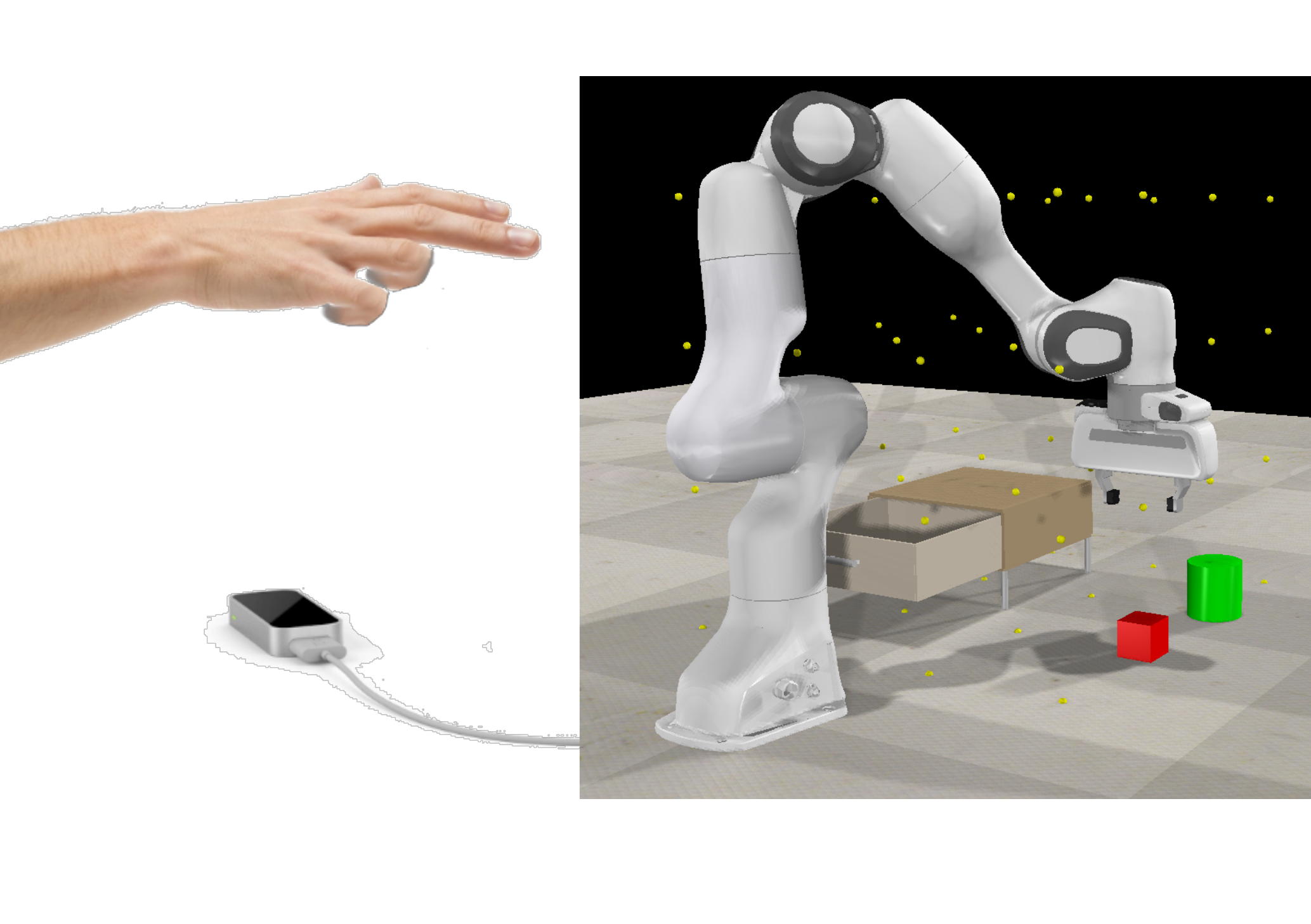}
  \caption{Experimental setup. On the left side is Leap Motion Sensor \cite{Weichert_Bachmann_Rudak_Fisseler_2013} for hand tracking. On the right side is a Robot setup with Panda Manipulator with an initialized scene. The yellow dots show the robot scene grid.}
  \label{fig:experimental_setup}
\end{figure}

We evaluate the proposed method using a simulated setup in the simulator Coppelia Sim \cite{rohmer_v-rep_2013} and a build tool PyRep \cite{james_pyrep_2019}. The scenes consist of a Franka Emika Panda with 7 DOF, cups, drawers, and cubes. The robot can open and close the drawer and manipulate the other objects. Furthermore, cups and cubes can be stacked on cubes and placed inside drawers. Cups also can contain liquid that can be poured into other containers. See the experimental setup in Fig.:~\ref{fig:experimental_setup}. 


\subsection{Gesture classification framework}
To classify the raw hand observations from users in the actual HRI case as gestures and compress them into the gesture vector, we utilize our \textit{Toolbox for gesture operation of the robot}~\cite{Vanc_Stepanova_Behrens}. The toolbox segments the episodes based on hand visibility. The classification is a signal per tracked gesture over time proportional to the probability that the gesture was presented at that point in time (see~Fig.~\ref{fig:episode_graph_01}).

\subsection{Artificial datasets generation}
We create artificial datasets with different levels of context-dependency of the gestures. These well-controlled datasets that simulate various users interacting with the system in different conditions enable us to evaluate which input data are necessary to consider to learn the mapping between the gestures and the intents. 

Each dataset comprises a set of observation points $\textbf{o}_i$:
\begin{equation}
\begin{split}
\textbf{D} &= \{\textbf{o}_i\},\\
\textbf{o}_i &= \left[\textbf{g}_j, \textbf{s}_i, \mathbf{f}_i, u_i \right],
\end{split}
\end{equation}
where \textbf{g} is a gesture observations vector, \textbf{s} is a scene description (object states, positions), \textbf{f} is a focus point, and $u$ is the identification of a user.  

Each observation is generated as follows: 1. a random scene generation, 2. one of the possible intent actions and target objects are selected, 3. a focus point is generated, 4. the user is selected, and 5. a gesture vector is generated. The individual steps are described in detail in the following subsections.

\subsubsection{Scene generation}
\label{sec:scene_generation}

\begin{figure}[htbp]
  \centering
  \includegraphics[width=0.4\textwidth]{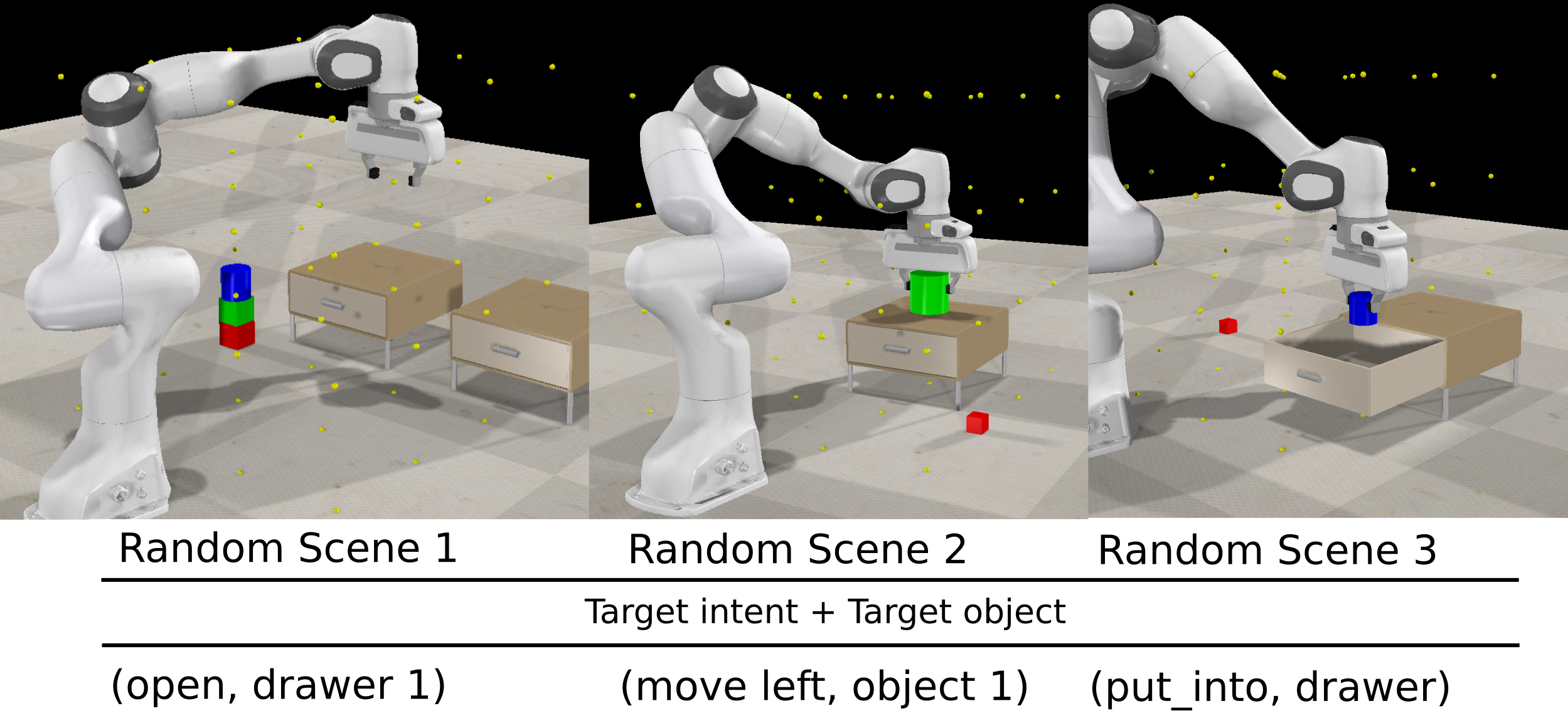}
  \caption{Random scene example generation.}
  \label{fig:cbgo_init_scene_random}
\end{figure}

To sample random valid scenes, we first randomly select the number $n$ of objects in the scene. To achieve useful combinations of scene objects, we instantiate the $n$ leading objects in the vector 
$$\mathbf{T} = \text{[cup, drawer, box, cup, drawer, box, cup]}.$$
Then we place each object randomly in a $4 \times 4 \times 4$ grid in front of the robot (all grid positions are reachable for the robot). First, drawers are placed, then cubes, and finally cups. Objects landing in the same vertical column are stacked, if possible, or replaced. Then each object pose is slightly perturbed. Finally, we randomly select the value of each object attribute, e.g., open/close drawer, full/empty cup, gripper full/empty. In the case of a full gripper, the pose of the assigned object is changed to the gripper location. Three sample generated random scenes are visualized in Fig.:~\ref{fig:cbgo_init_scene_random}.

\subsubsection{Generating Intents and Action sets}

For every scene, we generate a set of suitable intents. The set of valid intents is a subset of the Cartesian product of the scene objects $\times$ the target actions. $\text{Pre}(t_A)$ denotes the set of preconditions of the target action $t_A$ that must hold for execution. For invalid intents  holds
$$\bigcap \text{Pre}(t_A) \in \text{scene} \neq true.$$ 
Such intents are ignored. For example, the target action \textit{put into} is only applicable in scenes with an object in the gripper and an open drawer as the target object. 
\begin{equation}
\begin{split}
    \text{Pre}(t_A) =  \{&[\text{type}(t_O) = \textit{drawer}], \\ & [t_O.\text{open} = \textit{true}], \\ & [\text{gripper}.\text{holding} = \textit{true}]\}
\end{split}
\end{equation}


For \textit{pick up} only \textit{cubes} and \textit{cups} can be selected as target objects and the gripper has to be empty. In general, intents denote a desired delta to the current scene state. Note that we do not require that all robotic motions to reach the intent are specified. We assume that the robot will have a sufficient level of autonomy to work toward these tiny goals. 

The available intent action set in our setup includes \textit{[put into, put on target, place, pour, pick up, open, close, $\langle$ moves $\rangle$].}, moves:\textit{[right, left, up, down]}. One or more parameters refine each intent. For example, \textit{pour} is refined by an angle $\alpha$ that determines the final rotation of the object (default $\alpha=90\deg$).

\subsubsection{Generation of the focus point} Given the selected target object, the focus point ($f$) is selected from the normal distribution with the center in the target object. 
\subsubsection{Generation of the user} User $u$ is randomly selected.
\subsubsection{Generation of the gesture vector}
\label{sec:generation_of_the_observations}

The base for generating the gesture observations is a valid scene-intent-user combination. Simplified user models are used (inspired by a real human demonstrator performance within the given environment and the set of gestures). 
A multi-dimensional decision table represents the model. Each dimension represents the dependency on some context variable, such as the user, the scene, the intent, etc. For comparative studies, we prepared four datasets with a growing degree of differentiation in gesture generation. The four models (inspired by the selection of the gestures of a real demonstrator, see our website: \href{https://github.com/imitrob/context-based-gesture-operation}{github.com/imitrob/context-based-gesture-operation} for exact values) are of the following dimensions:
\begin{itemize}
    \item Dataset 1 $D1: [ I \times G ]$,
    \item Dataset 2 $D2: [ T \times I \times G ]$,
    \item Dataset 3 $D3: [ U \times T \times I \times G] $,
    \item Dataset 4 $D4: [ S \times U \times T \times I \times G]$,
\end{itemize}
where $G$ is the number of gestures in the sample set (in our case 9), $I$ is the number of intents in the sample set (in our case 11), $T$ is the number of object types (in our case 3), $U$ is the number of predefined users (in our case 2), and $S$ is the number of object states (in our case 2).

Higher-dimensional models capture that humans use gestures in a context-dependent way.
For example, one entry in Dataset 4 encodes
which gesture user 1 would use in a situation when the target intent \textit{put in} should be applied to the target object \textit{drawer} in the state \textit{open}. In this way, a set of different fake users was created.

\subsection{Compared models}
\label{sec:nn_compared_models}
To show how the mapping between the context-dependent gestures and the intent can be learned from observations, we compare the following five models of probabilistic neural networks (all stemming from the general scheme shown in Fig.~\ref{fig:model_graph_01}). Model M1 has one hidden layer. Other models have two hidden layers. The models differ in which input data they consider: \textit{M1} (gesture vector), \textit{M2} (gesture vector + dist.objects to focus point), \textit{M3} (gesture vector + user id), \textit{M4} (gesture vector + dist.objects to focus point+ user id), \textit{M5} (gesture vector + dist.objects to focus point + user id+ objects' states).

The models M1-M5 were trained on the datasets D1-D4 (see Sec.~\ref{sec:generation_of_the_observations}). Each dataset was split to training ($D_{train} = 4000$ samples) and testing ($D_{test} = 500$ samples) part. These were kept the same for all the evaluated models.


\section{Experimental Results}
\label{sec:expres}

This section describes an evaluation of each part of the process in the pipeline. 

\subsubsection{Gesture classification results}

\begin{figure}[htbp]
  \centering
\includegraphics[width=0.45\textwidth]{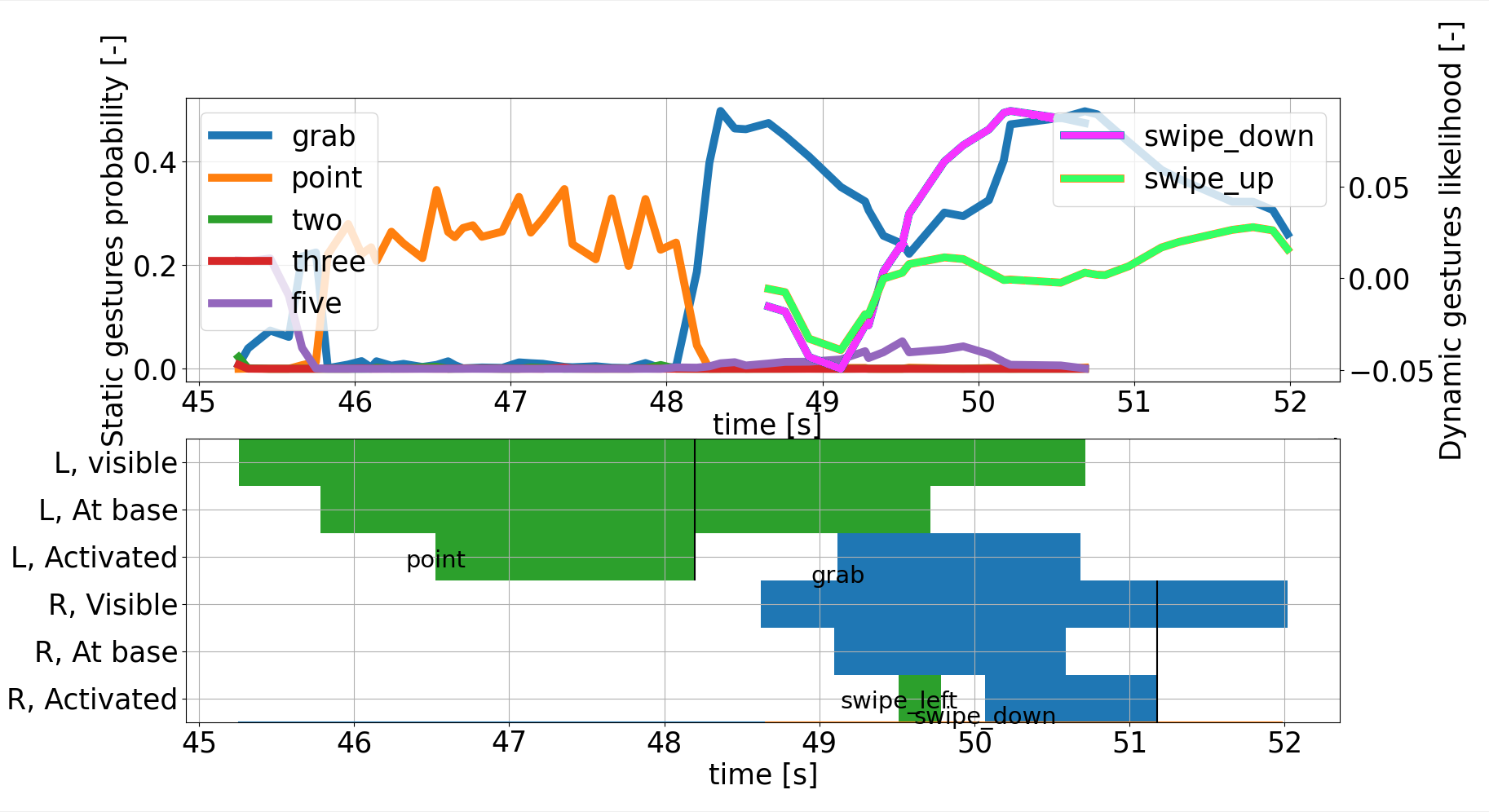}
  \caption{The gesture detection output of a single episode. The upper plot shows
the likelihoods for static (left legend) and dynamic (right legend)
gestures. The bottom plot shows the visibility of the hands and
the activation of the gestures. First was detected the static gesture
point followed by grab and dynamic gesture swipe down.
  }
  \label{fig:episode_graph_01}
\end{figure}

When the set of gestures is chosen correctly, the gesture feature space is not overlaying. Detection is unambiguous. For the static gesture set consisting of 8 gestures (grab, pinch, point, two, three, four, five, thumbs up), we reached $99.1\%$ balanced accuracy on the well-performed test data. The training dataset has around $12000$ samples and $4000$ for testing. The dynamic gesture detection approach based on the DTW was tested on a set of 5 directional swipe gestures and reached overall balanced accuracy of $83\%$.
Static and dynamic forms of gestures produce two independent streams of data. From one episode, we get a list of the few triggered gestures which fulfill the requirements. The reader can see single episode detection in Fig.:~\ref{fig:episode_graph_01}.

\subsubsection{Mapping results}

\begin{figure}[htbp]
  \centering
   \includegraphics[width=0.45\textwidth]{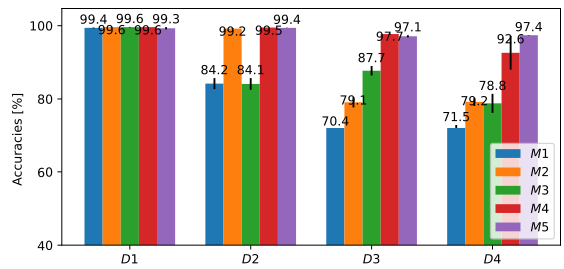}
  \caption{Model comparison trained on the generated datasets. Independent samples on the test set obtained the measured, balanced accuracies. Models are defined in Sec.:~\ref{sec:nn_compared_models} and datasets in Sec.:~\ref{sec:generation_of_the_observations}.
  }
  \label{fig:model_comparison}
\end{figure}

\begin{figure}[htbp]
  \centering
  \includegraphics[width=0.3\textwidth]{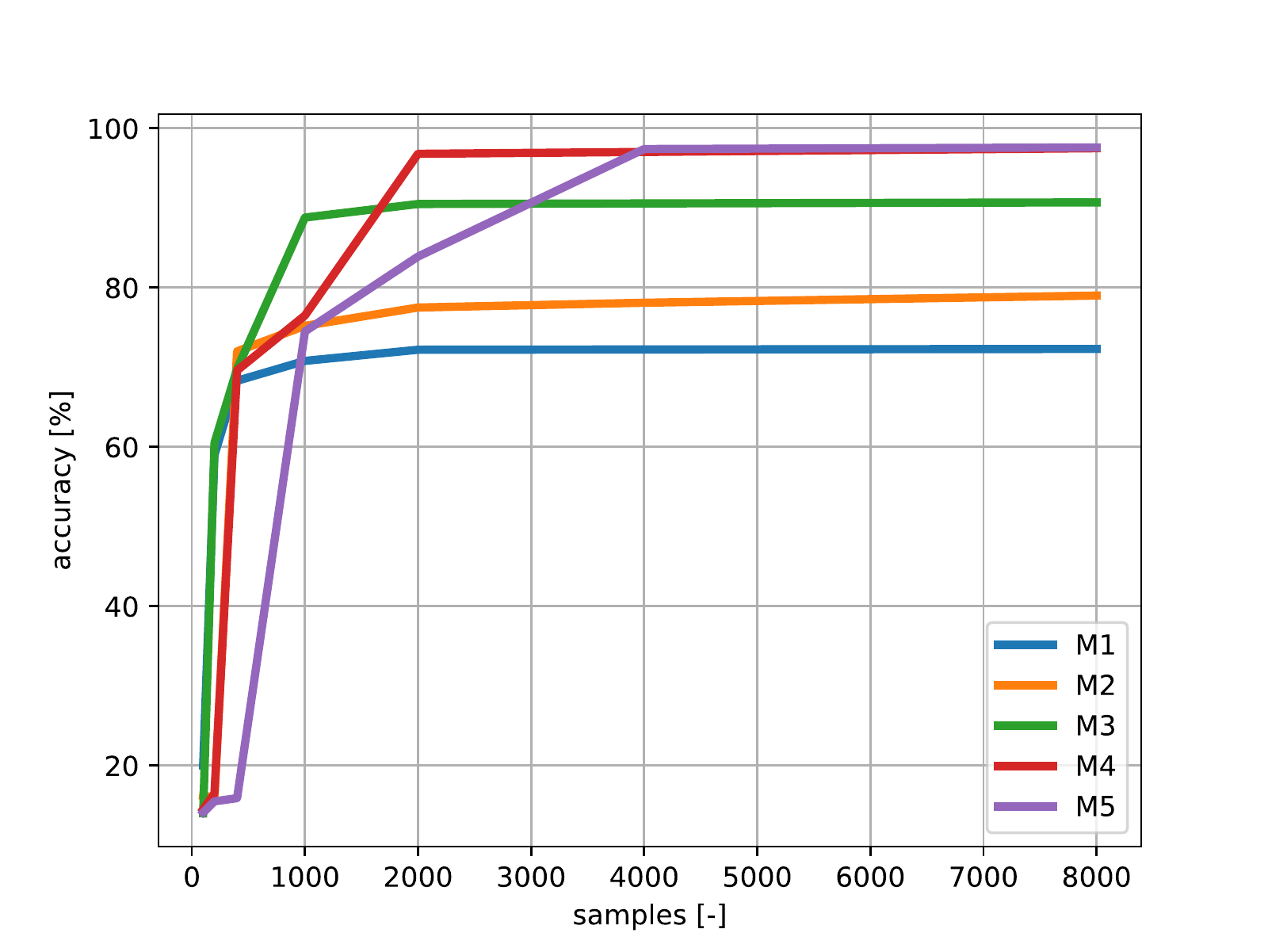}
  \caption{Accuracy of models trained on the dataset $D4$. }
  \label{fig:accuracy_on_samples}
\end{figure}

We evaluated the models $M1$-$M5$ (see Sec.~\ref{sec:nn_compared_models}) on datasets $D1$-$D4$ with gradually increasing context-dependency of the gestures (see Fig.:~\ref{fig:model_comparison}) using independent test dataset. The simple model considering only gesture vectors can learn the one-to-one mapping well in the context-independent dataset $D_1$. For datasets where gestures are used based on the context, models that consider more information about the scene and the user excel. If an sufficiently complex model is used, accuracies over $97\%$ are achieved on the testing set even for the most complex dataset. 
As a final experiment, the accuracy was measured concerning the number of samples needed for the network to learn the mapping for the dataset D4 well (see Fig.:~\ref{fig:accuracy_on_samples}). $2000$-$4000$ samples are needed to reach convergence. After $300$ samples, the accuracy improves significantly. In this case, we used the most context-dependent dataset $D4$, meaning that the system has to observe various situations to fully understand the gestures' context dependency. 


\section{Conclusion and discussion}
\label{sec:discussion}

In this paper, we proposed a system that uses functional gestures to operate robots. We show what the robot's intuitive, context-aware, and robust system for gesture operation might look like. This includes real-time detection of the gestures (modeled by probabilistic motion primitives) and the ability to extend the gesture set on the fly (using the Leap Motion sensor). Furthermore, such a system should be able to understand the human intent from the available observations of the hand movements (accumulating observations of the detected gestures over time) and the observed scene (e.g., objects on the scene and their state, state of the robot, user identification, etc.). Finally, we showed how to make the system robust and autonomous during the execution of the individual intents by utilizing behavior trees. 

To simulate the interaction of various users with the system, we prepared artificial datasets in a simulated robotic environment with different levels of context-dependency of the gestures. We show that incorporating the data about the context into the probabilistic neural network enables the models to learn the mapping between the gestures and the intents even when starting from zero knowledge about the mapping.


\section{Acknowledgment}
This work was supported by the European Regional Development Fund under project Robotics for Industry 4.0 (reg. no. $CZ.02.1.01/0.0/0.0/15\_003/0000470$), MPO TRIO project num. FV40319, and   by the Czech Science Foundation (project no. GA21-31000S).


\bibliographystyle{IEEEtran}

\bibliography{Gestures}

\end{document}